\documentclass[10pt,twocolumn,letterpaper]{article}

\usepackage{iccv}
\usepackage{times}
\usepackage{epsfig}
\usepackage{graphicx}
\usepackage{amsmath,bm}
\usepackage{amssymb}
\usepackage{multirow}
\usepackage{booktabs}

\DeclareMathAlphabet\mathzapf{T1}{pzc}{mb}{it}

\usepackage{stackengine}
\usepackage{enumitem}
\usepackage{comment}
\usepackage{caption} 
\usepackage[dvipsnames]{xcolor}
\usepackage[pagebackref=true,breaklinks=true,letterpaper=true,colorlinks,bookmarks=false]{hyperref}

\iccvfinalcopy 

\def\iccvPaperID{12738} 

\ificcvfinal\fi

\begin{document}

\title{ETran: Energy-Based Transferability Estimation}

\author{Mohsen Gholami$^{1,2}$, Mohammad Akbari$^{1}$, Xinglu Wang$^{1}$, Behnam Kamranian $^{1}$, Yong Zhang$^{1}$\\
$^{1}$Huawei Technologies Canada, $^{2}$University of British Columbia\\
{\tt\footnotesize \{mohsen.gholami,mohammad.akbari,xinglu.wang,behnam.kamranian,yong.zhang3\}@huawei.com}}

\maketitle
\ificcvfinal\thispagestyle{empty}\fi

\begin{abstract}
This paper addresses the problem of ranking pre-trained models for object detection and image classification. Selecting the best pre-trained model by fine-tuning is an expensive and time-consuming task. Previous works have proposed transferability estimation based on features extracted by the pre-trained models. We argue that quantifying whether the target dataset is in-distribution (IND) or out-of-distribution (OOD) for the pre-trained model is an important factor in the transferability estimation. To this end, we propose \emph{ETran}, an energy-based transferability assessment metric, which includes three scores: 1) energy score, 2) classification score, and 3) regression score. We use energy-based models to determine whether the target dataset is OOD or IND for the pre-trained model. In contrast to the prior works, \emph{ETran} is applicable to a wide range of tasks including classification, regression, and object detection (classification+regression). This is the first work that proposes transferability estimation for object detection task. Our extensive experiments on four benchmarks and two tasks show that \emph{ETran} outperforms previous works on object detection and classification benchmarks by an average of 21\% and 12\%, respectively, and achieves SOTA in transferability assessment. Code is available 
\href{https://developer.huaweicloud.com/develop/aigallery/notebook/detail?id=56e6645a-8133-49f6-a7ef-d877ef608fa5}{here\footnote{\url{https://developer.huaweicloud.com/develop/aigallery/notebook/detail?id=56e6645a-8133-49f6-a7ef-d877ef608fa5}}}.
\end{abstract}


\section{Introduction}
{

Pre-trained neural networks are widely available on platforms such as HuggingFace \cite{wolf2019huggingface} and TensorFlowHub \cite{tensorflow2015-whitepaper} for different tasks such as classification, object detection, segmentation, and natural language processing.
 
These pre-trained models, which have acquired the fundamental knowledge in vision \cite{kornblith} or language \cite{bert} domains, are very important in transfer learning to downstream target tasks with limited training data. 
Since training these models from the scratch are computationally expensive, task-specific fine-tuning from the pre-trained checkpoints is commonly considered as a time- and cost-efficient alternative solution.}

\begin{figure}[t]
\begin{center}
   \includegraphics[width=0.8\linewidth]{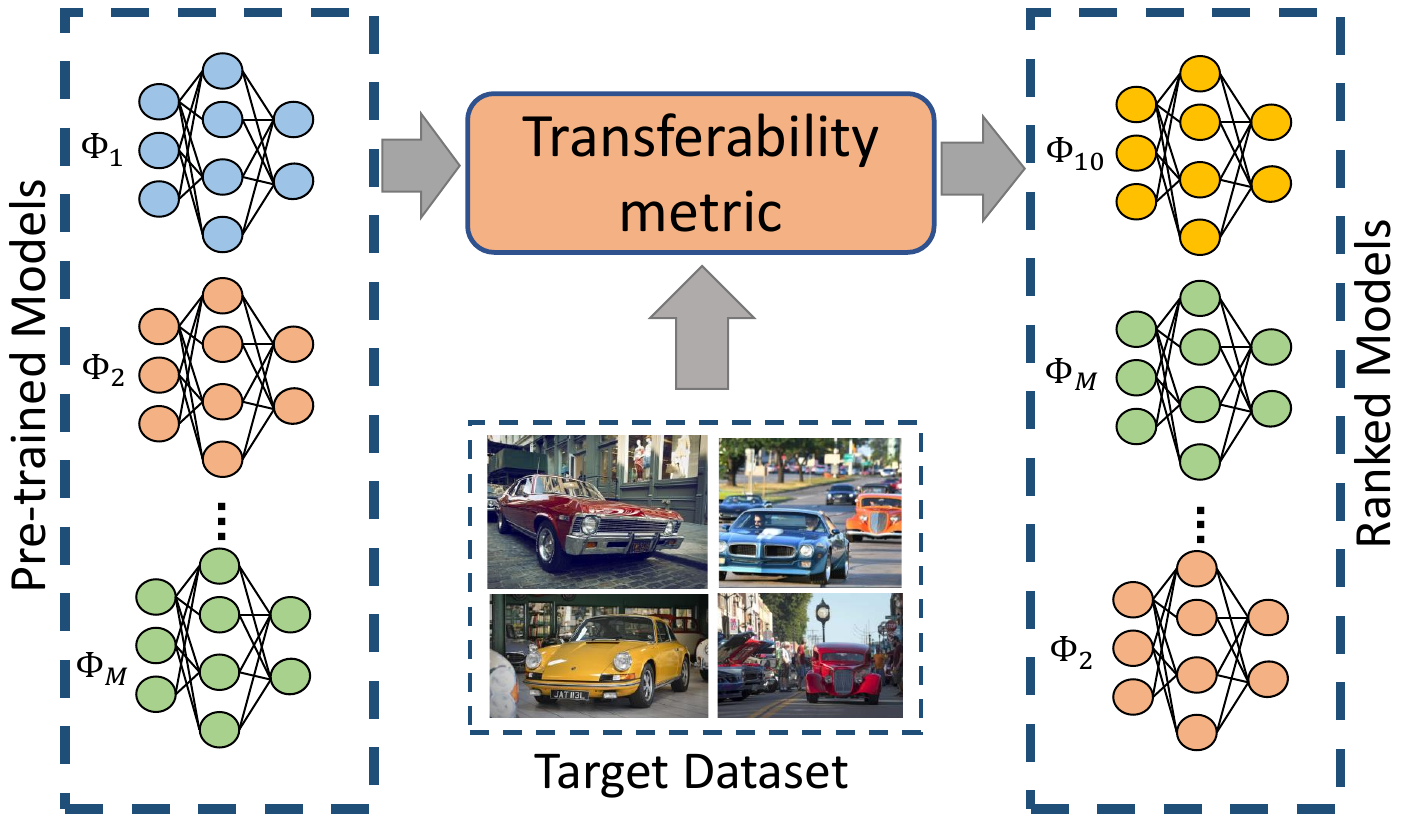}
\end{center}
    \caption{The overall framework of transferability assessment, given $M$ pre-trained models and a target dataset. }

\label{fig:tiser}
\end{figure}

{One of the major challenges in transfer learning is to select the best pre-trained model for a target task (or dataset), given numerous pre-trained models. The trivial solution to this problem is brute-force fine-tuning, where all the given pre-trained models need to be fine-tuned on the given dataset and then the best performing fine-tuned model is chosen for the target task. However, this procedure is very time-consuming and computationally expensive although it is highly accurate. Recent studies have proposed fast transferability assessment and ranking solutions to properly and quickly rank the models and select the best ones (Figure \ref{fig:tiser}).}

Most of the previous works extract features from the target dataset using the pre-trained models and try to find the model whose features can be more effectively mapped to the labels from the target dataset (e.g., $\mathcal{N}$LEEP {\cite{nleep}}, LogME {\cite{logme}}, PACTran {\cite{pactran}}, and SFDA {\cite{sfda}}). In other words, these methods try to mitigate the fine-tuning process  
in order to find a model that is more compatible to the given data samples based on the extracted features.

{
One limitation of these approaches is that the extracted features are drastically different before and after fine-tuning, which is due to the difference in source and target domains. As a result, the transferability assessment only based on the extracted features cannot lead to a reliable, general, and task-agnostic solution for obtaining an optimal pre-trained model. This uncertainty arises when the pre-trained model sees the input that differs from its training data (called in-distribution data), which can result in unreliable features and predictions \cite{oodenergy}. In other words, if the target dataset does not follow the data distribution with which the model has been trained, the extracted features cannot be reliable for transferability assessment. Thus, determining whether the target dataset is in-distribution (IND) or out-of-distribution (OOD) is an essential assessment factor in finding the best pre-trained model.
}

{
In this work, we propose an energy-empowered transferability estimation method (called \textit{ETran}) that exploits energy-based models (EBM) \cite{ebm} to detect whether a target dataset is IND or OOD for a given pre-trained model. To this end, the higher the energy score for a target dataset, the more IND this dataset is for the pre-trained model \cite{oodenergy,ebjr,elang}. As a consequence, the corresponding model has high likelihood to provide the best accuracy after fine-tuning (for the given target dataset) compared to the models with lower energy scores. In contrast to the previous transferability metrics, the energy score is label- and optimization-free, which makes it highly efficient and easy-to-use.}

{
Another major limitation of most of the previous works is that they are only applicable to classification tasks. H-Score \cite{hscore} and LogME \cite{logme} are the only works in the literature that introduce a solution for regression as well. Unlike the previous works, our method can deal with all classification, regression, and object detection tasks. To the best of our knowledge, \textit{ETran} is the first transferability assessment approach that is also applicable to object detection (i.e., a combination of classification and regression tasks).
}

\begin{figure*}[!t]
\begin{center}
   \includegraphics[width=0.99\linewidth]{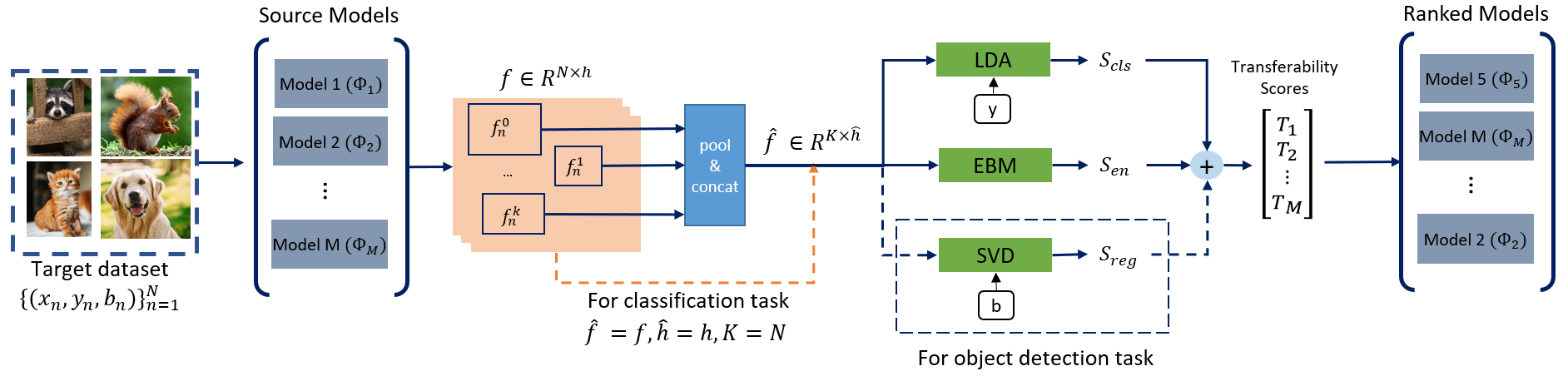}
\end{center}
    \caption{Overview of \textit{ETran}'s framework. $\Phi_m$: $m$th pre-trained source model, $f$: extracted features for the entire image, $f_{(n)}^{k}$: extracted features from $n$-th image and $k$-th bbox in the image (for object detection case), $S_\text{cls}$: LDA-based classification score, $S_\text{reg}$: SVD-based regression score, $S_\text{en}$: Energy-based score, $T_m$: the overall transferability score for the $m$th model.}
\label{fig:framework}
\end{figure*}

{
In addition to the energy score, we also propose to use classification and regression scores to benefit from the feature-based characteristics of general transferability metrics as in the previous methods. For the classification score, we use Linear Discriminate Analysis (LDA) \cite{hastie2009elements} to project features to a discriminate space by maximizing the variance between class labels and minimizing the variance within the classes. Bayes' theorem is then applied to measure the probability of class labels given the input features. For the regression score, we employ a solution based on Singular Value Decomposition (SVD) \cite{golub1971singular} that has fewer assumptions and better performance compared to LogME. Our experiments show that the regression score is crucial for an accurate transferability measurement on the object detection tasks. Figure \ref{fig:framework} shows the overall framework of \textit{ETran}.
}

The major {\bf contributions} of this work are as follows:
\begin{itemize}
    \item An energy-based transferability assessment metric that is label- and optimization-free; and also applicable to both classification and detection.
    \item Proposing two more scores based on LDA and SVD for transferability measurement on classification and object detection.    
    \item Proposing the first transferability metric for object detection tasks, and introducing multiple corresponding benchmarks and baselines that avail future research studies.
    \item Achieving SOTA results in transferability assessment for image classification and object detection.
\end{itemize}

\section{Related Work}
In this section, we discuss the transferability estimation methods introduced in the literature for both classification (including LEEP \cite{LEEP}, $\mathcal{N}$LEEP \cite{nleep}, PACTran \cite{pactran}, SFDA \cite{sfda}, and GBC \cite{gbc}) and regression (including H-Score \cite{hscore} and LogME \cite{logme}) tasks. 

LEEP \cite{LEEP} estimates the transferability of a source model to a target dataset by estimating two probabilities: 1) the predicted classes of the target dataset by the source model. 
LEEP used the prediction head of the source model which makes the transferability estimation limited to the source models trained in a supervised fashion for classification tasks. In contrast to LEEP, $\mathcal{N}$LEEP \cite{nleep}, PACTran \cite{pactran}, SFDA \cite{sfda}, and GBC \cite{gbc} use the features extracted from the target dataset by the source model to estimate the transferability. $\mathcal{N}$LEEP is indeed an extension of LEEP, which replaces the head detection of the source model with an Gaussian Mixture Model (GMM) fitted on the target data and then compute the LEEP score. PACTran \cite{pactran} argues that LEEP and $\mathcal{N}$LEEP overlook the generalization of source models and emphasize on the training error of the source dataset. Therefore, PACTran uses a linear model with a flatness regularizer to fit features to target labels using an optimization approach. 

SFDA \cite{sfda} and GBC \cite{gbc} propose that class separability of target dataset in the features space of source models is an important factor for transferability in classification scenarios. SFDA proposes to project features to a class separable space before applying a Bayes classifier. GBC uses Bhattacharyya coefficients to estimate the overlap between target classes in the features extracted by the source models. 

All of the above-mentioned works are exclusively applicable to classification tasks. 
These methods cannot be directly applied to the regression tasks due to the use of cross-entropy loss function \cite{pactran} or the class separating-based metrics \cite{gbc, sfda}.
On the other hand, H-Score \cite{hscore} and LogME \cite{logme} are the two methods in the literature that use least squares optimization for regression tasks. Their methods are also applicable to classification, where the problem is treated as a multi-variant regression problem. 

\section{Method}
\subsection{Problem Formulation}

\noindent{{\textbf{Classification.}}} Given $M$ pre-trained models $\{\Phi_{m}\}_{m=1}^{M}$ and a target dataset $\mathcal{D}=\{(x_{n},y_{n})\}_{n=1}^{N}$ ($N$: number of samples and $y$: ground-truth labels), the transferability metric for the $m$-th model is then computed as a scalar score $T_m$ as follows:
\begin{equation}
    T_{m}=\frac{1}{N}\sum_{n=1}^{N} p(y_{n}|x_{n},\Phi_m),
\end{equation}
where $p(y_{n}|x_{n},\Phi_m)$ is the probability of label $y_n$, given the input data $x_n$ and the source model $\Phi_m$. 

\noindent{{\textbf{Object Detection.}}} For the case of object detection, the target dataset is defined as  $\mathcal{D}=\{(x_{n},y_{n},b_{n})\}_{n=1}^{N}$, where $b_n$ denotes the bounding boxes (bboxes) labels in the $n$-th data sample. The transferability metric is also modified to  $T_{m}=\frac{1}{N}\sum_{n=1}^{N} p(y_{n},b_{n}|x_{n},\Phi_m)$.

\subsection{Feature Construction}
The transferability scores in our work are computed over the features extracted from the target dataset by the source models. In the classification scenario, the corresponding extracted features ($f$) are directly used as the input to the transferability metrics. However, preparing features for the object detection task is not straightforward due to the presence of bboxes in addition to the class labels. Using the entire feature vector $f$ for this task does not precisely provide bbox-specific features aligned with the ground-truth bboxes. In order to construct bbox-specific feature vectors, denoted by $f_{n}^k$, the ground-truth bboxes from the entire target dataset are utilized. To this end, $f_{n}^k$ represents the feature values in $f$ that exist in the relative position of the $k$-th bbox of the $n$-th sample. Since bboxes are of different sizes, adaptive average pooling is applied to map $f_{n}^k$ to a feature vector of size $\hat{h}$. All the pooled feature vectors are then concatenated to construct the overall feature vector $\hat{f}\in \mathbb{R}^{K\times \hat{h}}$, where $K$ is the total number of bboxes in the target dataset. The target feature dataset is then denoted by $\mathcal{F}:\{(\hat{f}_{k},y_{k},b_{k})\}_{k=1}^{K}$. In classification task  $\hat{f}=f$, $\hat{h}=h$, and $K=N$.

\subsection{ETran}
\label{ssec:ETran}
{\emph{ETran} is a hybrid transferability metric including energy, classification, and regression scores. The classification and regression scores are crucial since there is no single objective function which is optimal for both, especially in case of object detection. The energy score is also important as the other two scores are unable to determine whether the target dataset is IND or OOD for the pre-trained model.} The \textit{ETran}'s overall transferability metric is defined using the following score:
\begin{equation}
\label{eq:overall_score}
    T=S_\text{en}+S_\text{cls}+S_\text{reg},
\end{equation}
where $S_\text{en}$, $S_\text{cls}$, and $S_\text{reg}$ are the energy, classification, and regression (only for object detection) scores, respectively. 

\textbf{Energy Score.}
Energy-based models (EBM) introduce a function $E(x):\mathbb{R}^D \rightarrow \mathbb{R}$ that maps input data $x$ to a single, non-probabilistic scalar called \textit{energy} \cite{ebm}. Energies are uncalibrated values and can be turned into probability density $p(x)$ through \textit{Gibbs distribution}:
\begin{equation}
    p(y|x)=\frac{e^{-E(x,y)}}{\int_{y'}e^{-E(x,y')}},
    \label{eq:energy_f1}
\end{equation}
where the denominator $\int_{y'}e^{-E(x,y')}=e^{-E(x)}$ is called \textit{partition function}. The negative log of the partition function is Helmholtz free energy $E(x)$ of an input data $x$. The $p(y|x)$ obtained by EBM can also be calculated by a machine learning model $\Phi:\mathbb{R}^D\rightarrow \mathbb{R}^C$ by applying a softmax function as follows:
\begin{equation}
    p(y|x)=\frac{e^{\Phi^{(y)}(x)}}{\sum_c^{C}e^{\Phi^{(c)}(x)}},
    \label{eq:energy_f2}
\end{equation}
where $\Phi^{(y)}(x)$ is the output logits of $y$-th class. Due to the deep connection between EBMs and discriminative models, we can define the energy for a given data point $x$ as $E(x,y)=-\Phi^{(y)}(x)$ by equating the Eq. \ref{eq:energy_f2} and \ref{eq:energy_f1}. We can then compute the free energy $E(x)$ (defined as the negative log of partition function) as follows:
\begin{equation}
    E(x)=-\log\sum_c^C e^{\Phi^{(c)}(x)}.
\end{equation}

In the transferability estimation, we aim to assign low likelihood to the features extracted by a low-ranked source model and high likelihood to the features extracted from the high-ranked source models. If we take the logarithm of Eq. \ref{eq:energy_f1}, the following can be obtained:
\begin{equation}
\label{eq:energy_constz}
    \log p(x)=-E(x)-\underbrace{\log Z}_{\text{constant}},
\end{equation}
{where $Z$ is the denominator in Eq. \ref{eq:energy_f1}, which is constant for all samples.} Therefore, the negative free energy is correlated with the likelihood of samples. This means that samples with high energy values have less likelihood and are OOD samples for the source model $\Phi$. On the other hand, samples with high energy values are IND for $\Phi$. We hypothesis that $\Phi$ has a high transferability to a target dataset $\mathcal{D}$, if samples from $\mathcal{D}$ are in-distribution samples for $\Phi$. Therefore, the transferability score of $\Phi$ to $\mathcal{D}$ is correlated to free energy score of $\Phi$. 
In the above formulas, the free energy was calculated using output logits of $\Phi$. This has a major drawback as the logits are task-specific outputs which depend on the number of classes in the source dataset. On the contrary, features extracted by $\Phi$ are task-independent outputs that can be assumed as the output of the discriminative model $\Phi$ with $\hat{h}$ classes. Given $\hat{h}$ as the dimension of features $\hat{f}$, we calculate the energy values over features: \begin{equation}
    \hat{E}(x)=-\log\sum_{\eta}^{\hat{h}} e^{\hat{f}^{(\eta)}}.
\end{equation}
Having the energy values corresponding to all data samples $\{x_{k}\}_{k=1}^{K}$, we define the energy-based transferability score as follows:
\begin{equation}
    S_\text{en}=-\frac{1}{K}\sum_{k=1}^{K} \hat{E}(x_k).
\end{equation}

\textbf{LDA-Based Classification Score.}
Features extracted by the pre-trained models are separable based on the source dataset's classes. However, after fine-tuning, the features are separable based on the target dataset's classes.
The classification score in this work is obtained using Bayes theorem after projecting the features into a subspace that separates features of different target classes as much as possible. 

In this work, the projection matrix, denoted by $U$, is computed using Linear Discriminant Analysis (LDA) \cite{hastie2009elements}:
\begin{equation}
    U=\text{arg} \stackunder{ max }{U}\frac{U^{T}\Sigma_{\beta}U}{U^{T}\Sigma_{\omega}U},
    \label{eq:proj}
\end{equation}
where $\Sigma_{\beta}=\sum_{c=1}^{C}K_c(\mu_c-\mu)(\mu_c-\mu)^{T}$ and $\Sigma_{\omega}=\sum_{c=1}^{C}\sum_{k=1}^{K_c}(\hat{f}^{(c)}_{k}-\mu)(\hat{f}^{(c)}_{k}-\mu)^{T}$ are the between-scatter and within-scatter matrices, respectively. $\mu_c$ and $\mu$ are the mean of $c$-th class and the total mean of the target data. $C$ is the number of classes in the target dataset and $K_c$ is the number of samples (i.e., bboxes in object detection) in $c$-th class. $\hat{f}_{k}^{(c)}$, which is obtained by splitting $\mathcal{F}$ into classes, represents the feature vector corresponding to the $k$-th sample (or bbox) with $c$-th class.
The optimization in Eq. \ref{eq:proj} is equivalent to \cite{ghojogh2019fisher} :
\begin{equation}
    \begin{split}
    \stackunder{\text{maximize  }}{U} U^T\Sigma_{\beta}U,
    \text{~~~~~subject to:  } U^T \Sigma_{\omega} U=1.
    \end{split}
\end{equation}
The Lagrangian of this optimization is defined as: 
\begin{equation}
    \mathcal{L}=U^T\Sigma_{\beta}U-\lambda(U^T\Sigma_{\omega}U-1),
\end{equation}
where $\lambda$ is the Lagrangian multiplier. Equating the derivative of $\mathcal{L}$ to zeros gives:
\begin{equation}
    \begin{split}
    \frac{\partial{\mathcal{L}}}{\partial{U}}= 2 \Sigma_{\beta}U-2\lambda\Sigma_{\omega}U=0 ~~~\Rightarrow~~~
    \Sigma_{\beta}U=\lambda \Sigma_{\omega}U.
    \end{split}
\end{equation}
The above eigenvalue problem can be solved as follows \cite{ghojogh2019eigenvalue}: 
\begin{equation}
    U=eig((\Sigma_{\omega}+\epsilon I)^{-1}\Sigma_{\beta}),
\end{equation}
where $\epsilon$ is a small positive number to make the $\Sigma_w$ full-rank in case that $\Sigma_w$ is singular. Having the projection matrix, $U$, we project the features by $\bar{f}=U^T\hat{f}$. We assume that each class has a normal distribution $\bar{f}^{(c)}\sim \mathcal{N}(U^T\mu_c,I)$ where $I$ is an identity matrix. 
Therefore, the Bayes theorem can be applied to obtain the prediction score $\delta_c$ for each class $c$.
The \textit{ETran}'s classification score is then defined as the probability of the ground-truth class as follows:
\begin{equation}
    S_{cls}=\frac{1}{K}\sum_{k=1}^{K}\frac{e^{\delta_y}}{\sum_{c=1}^{C}e^{\delta_c}},
    \label{eq:L_c}
\end{equation}
where $y$ is the ground-truth labels. 

\textbf{SVD-Based Regression Score.} 
Singular Value Decomposition (SVD) can be utilized for approximately solving linear regression in a way that it is less sensitive to errors and more effective for ill-conditioned matrices \cite{hansen1998rank}. This is due to the singular values in the diagonal matrix being sorted in descending order, so the smallest values can be truncated or set to zero without significantly affecting the overall solution. 
Therefore, we propose to use reduced SVD \cite{svd-reg,golub1971singular} to efficiently estimate the transferability of features obtained by the source model. We decompose the feature matrix $\hat{f}=U \text{diag}{(s)}V^{T}$ where $U \in\mathbb{R}^{K \times \hat{h}}$, $s\in \mathbb{R}^{\hat{h}}$, and $V \in \mathbb{R}^{\hat{h} \times \hat{h}}$. We then use the decomposed matrices to calculate the approximated pseudo-inverse \cite{golub1965calculating} of the features as follows: 
\begin{equation}
    \hat{f}^{\dagger}=V\text{diag}(\hat{s})^{-1}U^{T},
    \label{eq:svd}
\end{equation}
where $\hat{s}$ is the truncated singular values whose top 80\% is preserved.  
Given the bbox position labels $b \in \mathbb{R}^{K \times 4}$, we calculate the projection of $b$ into the subspace spanned by columns of $\hat{f}^{\dagger}$ by $\hat{f}^{\dagger}b$. The approximated labels $\hat{b}$ are calculated by $\hat{b}=\hat{f} \hat{f}^{\dagger} b$.

The \textit{ETran}'s regression score is then computed by the mean squared error between the approximated labels and ground-truth bbox labels as follows:
\begin{equation}
    S_{r}=-\frac{1}{K\times4}\sum_{k=1}^{K}{\sum^{4}_{j=1} (b_{k}^{(j)}-\hat{b}_{k}^{(j)})^{2}},
    \label{eq:reg_score}
\end{equation}
where $b_{k}^{(j)}$ denotes the $j$-th position value in the $k$-th bbox.

\subsection{Baselines for Object Detection}
\label{ssec:baselines}
In this section, we define 3 new baseline transferability metrics for object detection based on the SOTA methods by LogME \cite{logme}, PACTran \cite{pactran}, and SFDA \cite{sfda}, which have been originally proposed for classification tasks. These classification-based metrics can be directly applied to object detection by only evaluating the compatibility between bbox features and their class labels. However, such strategy does not consider the bbox information (as a regression problem) that is a crucial part of object detection tasks, which is required to achieve good performance in transferability assessment. In order to benefit from the bbox information in all of the 3 baselines in this work, we employ LogME's regression solution \cite{logme} to calculate our baseline regression score (denoted by $S_{\text{lmr}}$) as:
\begin{equation}
 \begin{split}
     S_{\text{lmr}} & = \frac{1}{4}\sum^{4}_{j=1} \big( \frac{1}{2}\log\gamma+\frac{\hat{h}}{2}\log\alpha-\frac{K}{2}\log2\pi \\
     &-\frac{\gamma}{2}||\hat{f}q-b^{(j)}||-\frac{\alpha}{2}q^{T}q-\frac{1}{2}\log||A|| \big),
 \end{split}
 \label{eq:LK}
 \end{equation}
where $A= \alpha I + \gamma \hat{f}^{T}\hat{f}$ and $q=\gamma A^{-1} \hat{f}^{T} b^{(j)}$. 
Here, $\alpha$ and $\gamma$ are positive parameters in the prior distribution of weights and observations. The weights map the features to target labels. 
As in Eq. \ref{eq:svd}, $b^{(j)}$ represents the $j$th position values, but for all $K$ bboxes.

The overall transferability metric for our baselines is then computed as follows:
$T = S_\text{baseline} + S_{\text{lmr}}$,
where $S_\text{baseline}$ is the classification score calculated via LogME, PACTran, or SFDA methods.

\section{Experiments}
In this section, the performance of the proposed transferability assessment method (\textit{ETran}) compared with the previous works is numerically evaluated on image classification and object detection. An extensive set of experiments over different benchmarks along with ablation studies and computational complexity analysis are also presented. 
In addition, three transferability assessment benchmarks based on VOC, COCO, and HuggingFace datasets are introduced for the object detection task.

\textbf{Evaluation Metric.} In order to evaluate the performance of the proposed method, the ground-truth ranking scores
of all the pre-trained models ($\Phi_{m}$) are required. The corresponding ranking scores, denoted by $G_{m}$, are basically the validation accuracies obtained after fine-tuning each $\Phi_{m}$ on the target dataset. Following the previous works \cite{pactran,gbc,sfda,logme}, we use Kendall's tau, denoted by $\tau$, as our main evaluation metric. Kendall's tau \cite{kendall} is defined as the number of concordant pairs minus the number of discordant pairs divided by the overall number of pairs $\binom{M}{2}$ as follows:
\begin{equation}
    \tau=\frac{2}{M(M-1)}\sum_{i=1}^{M}\sum_{j=i+1}^{M} \text{sgn}(G_{i}-G_{j})\cdot \text{sgn}(T_{i}-T_{j}).
\end{equation}
{We use the weighted version of Kendall's tau by \cite{weightedkendaltau}, $\tau_{w}$, that assigns more weights to the top ranked models. 
In addition to $\tau_{w}$, we also use the probability of correctly finding the top-k pre-trained models, denoted by $Pr(\text{topk})$, as another evaluation metric. $Pr(\text{topk})$ is the probability of the ground-truth top-ranked model being among the top $k$ estimated models. In this work, we report $Pr(\text{top1})$, $Pr(\text{top2})$, and $Pr(\text{top3})$.  
Although $\tau_w$ shows whether the whole ranking of the pre-trained models matches the ground-truth ranking, $Pr(\text{top-k})$ is also important in the real-case scenarios, where we only need to find the best pre-trained model.}

\begin{table*}\setlength\tabcolsep{5pt}
\footnotesize
\centering
\caption{Classification Benchmark: Performance (Kendall tau $\tau_w$) of different methods} 
\label{tab:Classification}
\begin{tabular}{l|ccccccccccc|c}
	\toprule
	Method             &  CIFAR10   &     VOC     &   Caltech-101   &  AirCraft  & CIFAR100  & Food-101  &  Pets      &Flowers    &Cars &DTD   &Sun & \textbf{Average}\\ \midrule
	LEEP \cite{LEEP}               &   0.824    &    0.413    &    0.605    &   -0.233   & 0.667 & 0.434     & 0.389 &  -0.242  &0.317&0.417&0.697&0.390 \\ 
	NLEEP \cite{nleep} & -0.360 & -0.233 & 0.281 & 0.332 & 0.696 & 0.468 & 0.230 & -0.162 & 0.367 & 0.378 & 0.511 & 0.228\\ 
	OTCE \cite{otce} &0.562       & 0.639            &           0.104  & 0.099      &   0.285   &     0.474  &  0.056&0.265&0.439&0.082&-0.139&0.260 \\ 
	LogME \cite{logme}             &   0.852    &    0.564    &    0.352    &   0.334    & 0.725 & 0.385 & 0.411 & -0.008  &0.485&0.662&0.545&0.482\\
	PACTran \cite{pactran} &0.562& -0.235 &0.528&-0.038&  0.763     & 0.000      & 0.318      &0.329  &-0.121&0.522&0.301&0.266 \\
	SFDA \cite{sfda} &  {0.849}   &    0.518    &   {0.555}   &   -0.215   & 0.793 & 0.427  & 0.340 & 0.590 &0.312&0.633&0.722&0.502 \\

	\hline
	LEEP+$S_\text{en}$ &0.897&0.413&0.626&-0.077&0.697&0.434&0.389&-0.070&0.405&0.417&0.658& 0.435\\
	LogME+$S_\text{en}$     & {\bf 0.890} &  0.656 & {\bf 0.567} & {\bf 0.370} &  0.774     & 0.484      &  0.447     & -0.021 &{\bf0.586}&{\bf0.682}&0.570& 0.545\\
	PACTran+$S_\text{en}$ &0.562	&-0.235	&0.528	&0.046	&0.702	&0.024	&0.437	&0.329	&-0.163	&0.599	&0.378&0.291
 \\ 
	SFDA+$S_\text{en}$      &    {\bf 0.890}    &    0.606    &    0.558    &   -0.161   &   0.856    &  0.370     &   0.422    &0.406  &0.328&0.639&{\bf0.744}&0.514 \\ \hline
	ETran ($S_\text{en}$)             &   0.816    &    0.476    &    0.41     &   0.331    & 0.557 & 0.396 & 0.307  &  0.277 &0.500&0.606&0.556&0.475\\
	ETran ($S_\text{cls}\text{+}S_\text{en}$)       &   0.887    &   \textbf{0.667}    &    0.440     &   -0.091   &   {\bf 0.900}   &  {\bf0.829}     &  {\bf0.713}     & {\bf0.580} &0.246&0.303&0.708&\textbf{0.562} \\
 \bottomrule
\end{tabular}
\end{table*}

\begin{figure*}[t]
\begin{center}
   \includegraphics[width=1\linewidth]{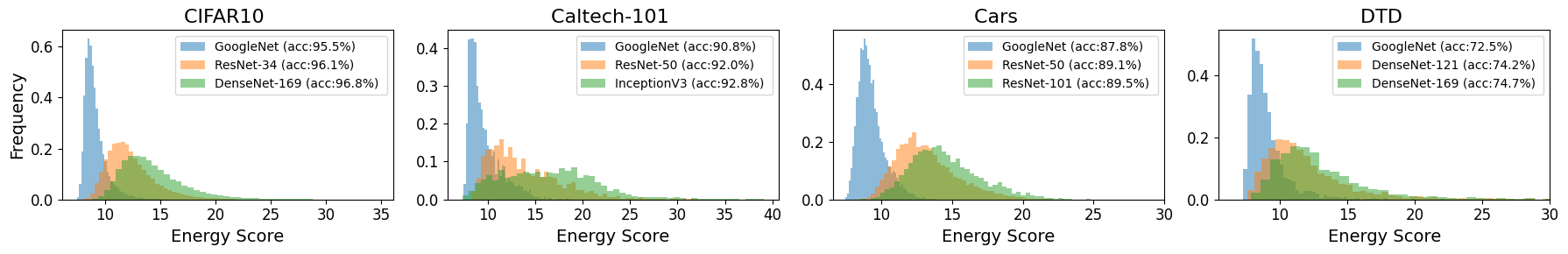}
\end{center}
   \caption{Energy score distributions corresponding to three source models on the CIFAR10, Caltech101, Cars, and DTD target datasets in the classification benchmark.}
\label{fig:energy_good}
\end{figure*}

\begin{figure}[t]
\begin{center}
   \includegraphics[width=1\linewidth]{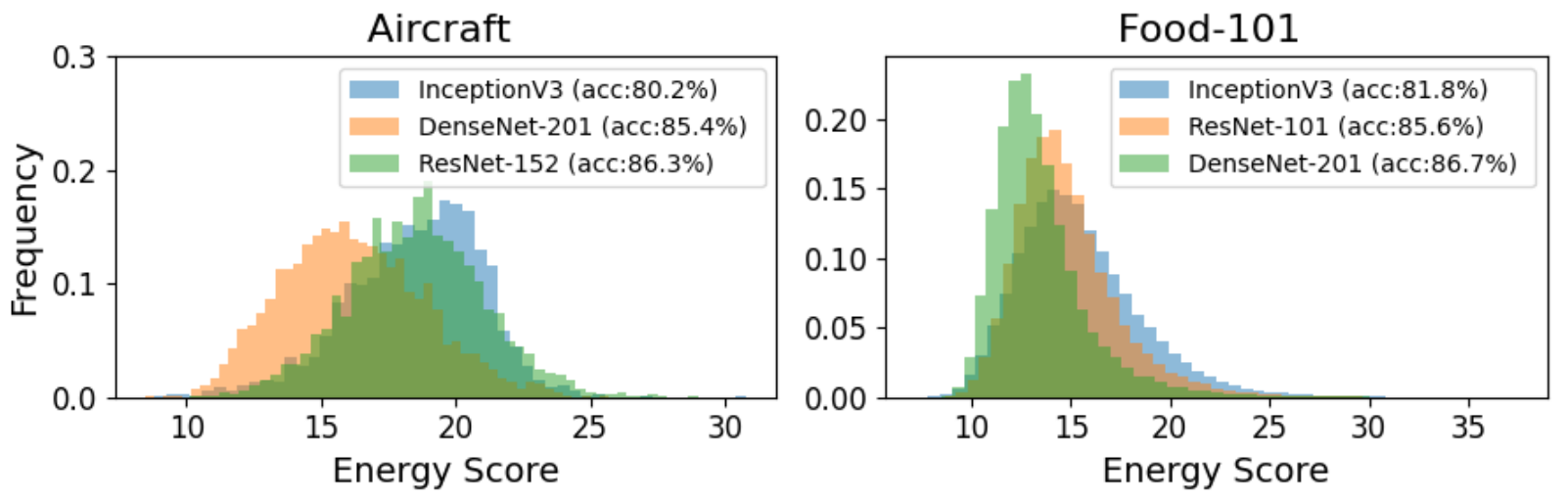}
\end{center}
   \caption{Some failure cases of ranking source models by energy scores.}

\label{fig:energy_bad}

\end{figure}

\subsection{Image Classification}
\label{ssec:experiments_cls}
\textbf{Benchmark.} 
The experiments for the classification task are performed on the benchmark used in \cite{sfda} that has 11 different source models (pre-trained on ImageNet) and 11 target datasets. The source models include ResNet-34, ResNet-50, ResNet-101, ResNet-152 \cite{resnet}, DenseNet-121, DenseNet-169, DenseNet-201 \cite{densenet}, MNet-A1 \cite{mnet}, MobileNetV2 \cite{mobilenet}, GoogleNet \cite{googlenet}, and InceptionV3 \cite{inceptionv3}. The target datasets include FGVC Aircraft \cite{aircraft}, Caltech-101 \cite{caltech}, Stanford Cars \cite{cars}, CIFAR-10, CIFAR-100 \cite{cifar}, DTD \cite{dtd}, Oxford-102 Flowers \cite{flowers}, Food-101 \cite{food}, Oxford-IIIT Pets \cite{pets}, SUN397 \cite{sun}, and VOC2007 \cite{voc2007}. We fine-tune all the source models on all of the target datasets to obtain the ground-truth scores, $G$ (details in the appendix).

{\bf Results Analysis.} 
{The \textit{ETran}'s transferability score for the classification scenario is defined as $T=S_\text{cls}$+$S_\text{en}$, where the regression score does not exist.
Table \ref{tab:Classification} demonstrates the results of \textit{ETran} compared with the previous works on the classification benchmark \cite{sfda}. \textit{ETran} outperforms all the previous works and achieves SOTA results with an average $\tau_w$ of 0.562 that is relatively 12\% better than SFDA.}

In order to show the effectiveness of the proposed energy score ($S_\text{en}$), we also integrated this score with all the previous works and report the results in Table \ref{tab:Classification}. It is shown that the energy score provides relative improvements for LEEP, LogME, PACTran, and SFDA by about 11\%, 13\%, 9\%, and 2\%, respectively, in terms of the average $\tau_w$.
Given the efficiency of the energy score calculation (i.e., almost 10$\times$ faster than the previous works), the corresponding improvements comes with a low cost.

{It is also shown that \textit{ETran}'s performance with only the energy score can obtain an average $\tau_w$ of 0.475, which is comparable with most of the previous works and even better than LEEP and PACTran. Since the energy score is completely unsupervised (no need for labels), our proposed method can be applied in the case that labels are not provided with the target datasets. It is an important merit in realcase scenarios, especially with costly labeling procedures.}

{\bf Energy Analysis.} 
{
Figure \ref{fig:energy_good} shows the energy score distributions corresponding to three source models on the CIFAR10, Caltech101, Cars, and DTD target datasets in the classification benchmark. The ground-truth validation accuracy of the models are also provided in the legends of the figures. We can observe that 
the source models with a higher accuracy on the target dataset have higher range of energy scores. For instance, on CIFAR10, Densnet169, ResNet-34, and GoogleNet are ranked as the first, second, and third models, respectively, which follow the same ranking in terms of having higher range of energy scores. Although it is the case for majority of the datasets, there are some rare cases, where models with higher accuracy give lower energy scores. Two examples are given in Figure \ref{fig:energy_bad}.}

\begin{table*}
\centering
\caption{Results on VOC-FT and VOC-RH object detection benchmarks.}
\footnotesize
\label{tab:VOC}
\begin{tabular}{ l|c|cccc|cccc}
\toprule
\multicolumn{2}{c|}{}&\multicolumn{4}{c|}{{VOC-FT}}& \multicolumn{4}{c}{{VOC-RH}}\\ \midrule
Method          &reg & Pr(top1)&  Pr(top2)&Pr(top3)&{$\bm\tau_{w}$}& Pr(top1)& Pr(top2)& Pr(top3) &{$\bm\tau_{w}$}\\ \hline
LogME \cite{logme}           &         &0.071&0.107&0.250&0.180&0.107&0.250&0.393&0.340\\
PACTran \cite{pactran}  &          &0.143&0.214&0.321&0.131&0.143&0.286&0.357&0.242\\
Linear  \cite{pactran} &          &0.143&0.214&0.321&0.132&0.143&0.286&0.357&0.242\\
SFDA \cite{sfda} &          &0.107&0.107&0.250&0.108&0.250&0.321&0.357&0.376\\\hline
LogME+$S_\text{lmr}$    &\checkmark&0.071&0.179&0.357&0.350&0.321&0.536&0.571&0.537\\
PACtran+$S_\text{lmr}$  &\checkmark&0.071&0.179&0.321&0.355&0.393&0.500&0.571&0.560\\
Linear+$S_\text{lmr}$&\checkmark&0.071&0.179&0.321&0.359&0.357&0.536&0.571&0.549\\
SFDA+$S_\text{lmr}$     &\checkmark&0.107&0.179&0.321&0.354&0.357&0.536&0.571&0.551\\
LogME+$S_\text{reg}$    &\checkmark&0.036&0.107&0.214&0.336&0.321&0.500&0.643&0.560\\
PACTran+$S_\text{reg}$  &\checkmark&0.071&0.107&0.321&0.335&0.214&0.429&0.571&0.508\\
Linear+$S_\text{reg}$ &\checkmark&0.071&0.143&0.393&0.352&0.250&0.429&0.607&0.555\\
SFDA+$S_\text{reg}$     &\checkmark&0.107&0.214&0.357&0.353&0.250&0.393&0.500&0.529\\ 
SFDA+$S_\text{reg}\text{+}S_\text{en}$ & \checkmark&0.214&0.321&0.536&0.462&0.143&0.393&0.500&0.528\\ \hline
\textbf{ETran ($S_\text{en}$)} &&{\bf0.286}&\textbf{0.393}&0.464&0.309&0.000&0.250&0.429&0.318
\\ 
\textbf{ETran ($S_\text{cls}\text{+}S_\text{en}\text{+}S_\text{reg}$)} &\checkmark&0.250&0.321&\textbf{0.536}&\textbf{0.464}&\textbf{0.500}&\textbf{0.536}&\textbf{0.679}& \textbf{0.590}\\ \bottomrule
\end{tabular}
\end{table*}
\begin{table}\setlength\tabcolsep{4pt}
\centering
\caption{Results on COCO object detection benchmark.}
\footnotesize
\label{tab:coco}
\begin{tabular}{ l|c|cccc}
\toprule
Method & reg & Pr(top1)&  Pr(top2) & Pr(top3) & {$\bm\tau_{w}$} \\\hline
LogME \cite{logme} &&0.267&0.400&\bf{0.600}&\underline{0.269}\\
PACTran \cite{pactran} &&0.133&0.267&0.333&0.138\\
Linear \cite{pactran} &&0.133&0.267&0.333&0.139\\
SFDA \cite{sfda} &&0.200&0.267&0.533&0.104\\\hline
LogME+$S_\text{lmr}$ & \checkmark&0.067&0.467&0.533&0.249\\
PACtran+$S_\text{lmr}$ &\checkmark&0.200&0.333&0.467&0.229\\
Linear+$S_\text{lmr}$ &\checkmark&0.200&0.333&0.467&0.227\\
SFDA+$S_\text{lmr}$ &\checkmark&0.200&0.333&0.467&0.183\\ \hline
\textbf{ETran ($S_\text{en}$)}&&\underline{0.267}&\underline{0.467}&{\bf0.600}&0.213
\\ 
\textbf{ETran ($S_\text{cls}\text{+}S_\text{en}\text{+}S_\text{reg}$)}  &\checkmark&{\bf0.400}&\bf{0.533}&{\bf0.600}&{\bf0.333}\\ \bottomrule
\end{tabular}
\end{table}
\begin{table}\setlength\tabcolsep{4pt}
\centering
\caption{Results on HF object detection benchmark.}
\footnotesize
\label{tab:HugFace}
\begin{tabular}{ l|c|cccc}
\toprule
Method&reg & Pr(top1)&  Pr(top2)&Pr(top3)&{$\bm\tau_{w}$} \\\hline
LogME \cite{logme}&&\bf{0.600}&0.600&0.800&0.374\\
PACTran \cite{pactran}&&0.400&0.400&0.600&0.140\\
Linear \cite{pactran}&&0.200&0.400&0.600&0.214\\
SFDA \cite{sfda}&&0.400&0.600&\bf{0.100}&0.312\\\hline
LogME+$S_\text{lmr}$&\checkmark&\bf{0.600}&0.600&0.800&0.400\\
PACtran+$S_\text{lmr}$&\checkmark&0.200&0.400&0.800&0.322\\
Linear+$S_\text{lmr}$&\checkmark&0.200&0.200&0.800&0.306\\
SFDA+$S_\text{lmr}$&\checkmark&0.200&0.400&0.800&0.202\\ \hline
\textbf{ETran ($S_\text{en}$)}&\checkmark&\underline{0.400}&\bf{0.800}&0.800&0.412\\
\textbf{ETran ($S_\text{cls}\text{+}S_\text{en}\text{+}S_\text{reg}$)}&\checkmark&\bf{0.600}&{\bf0.800}&{\bf1.000}&{\bf0.522}\\ \bottomrule
\end{tabular}
\end{table}

\subsection{Object Detection}

\subsubsection{Benchmarks and Setup}

{For the numerical analysis of our proposed transferablility metric for object detection, we design 3 benchmarks based on VOC2012 \cite{pascal-VOC-2012}, COCO \cite{coco}, and HuggingFace (HF) \cite{wolf2019huggingface} datasets.}

\textbf{VOC.} We split the VOC2012 dataset into two clusters called the source and target clusters. VOC2012 has 20 classes from which we assign 12 classes to the source cluster and 8 classes to the target cluster. We randomly select 3 classes from the source cluster and repeat this selection 19 times to create 19 different source datasets. The YOLOv5s object detection model \cite{yolo,yolov5} is trained on each of the source datasets resulting in 19 pre-trained source models. We also randomly select 28 pairs from the target cluster to create 28 target datasets (divided into train and validation sets). All pre-trained models are fine-tuned on the created target datasets (train set) and the \textit{map50} value over the validation sets is used to obtain the ground-truth ranking scores of pre-trained models.

Following the method in LEEP \cite{LEEP}, we use two approaches to fine-tune the source models: 1) fine-tuning the entire model, i.e., all layers (denoted by \textit{FT}), 2) re-training only the detection head from the scratch with all the other layers frozen, denoted by \textit{RH} (more details in the appnedix).

\textbf{COCO.} We apply the same above-mentioned procedure in VOC to the COCO dataset (with 80 classes in total). We consider 65 and 15 classes for the source and target clusters, respectively. 9 source datasets each with 20 classes randomly selected from the source cluster are created. 15 target datasets each with 2 classes randomly selected from the target cluster are also created.
Similar pre-training and fine-tuning process (only the \textit{FT} case) as in VOC is performed for this benchmark.

\textbf{HuggingFace (HF).} In the VOC and COCO benchmarks, the architecture of the source models was fixed, but it was trained on different source datasets. In this benchmark, we fix the source dataset, but use different model architectures. Six different models including YOLOv5s, YOLOv5m, YOLOv5n \cite{yolov5}, YOLOv8s, YOLOv8m, and YOLOv8n \cite{yolov8} are employed all of which are pre-trained on COCO dataset. We fine-tune these source models (all the layers) on 5 object detection datasets presented in the HuggingFace platform: Blood \cite{blood}, NFL \cite{nfl}, Valorant Video Game \cite{valorant}, CSGO Video Game \cite{csgo}, and Forklift \cite{forklift}. Note that the object classes in these target datasets have not been defined in COCO dataset. Therefore, the pre-trained models have not seen the targets classes beforehand.

\subsubsection{Results Analysis}

{\bf VOC.} 
{
The experimental results of \textit{ETran} on the VOC-FT and VOC-RH object detection benchmarks are summarized in Table \ref{tab:VOC}. 
In both scenarios, the source models are the same, however, the ground-truth rankings are different. It is worth mentioning that the VOC-FT case provides a more challenging transferability estimation because the features extracted by the source and target models are quite different.}

{
The baseline results with both LogME ($S_\text{lmr}$) and our SVD-based regression ($S_\text{reg}$) scores are given in Table \ref{tab:VOC}, which outperforms classification-only metrics in the previous works. For example, \textit{SFDA+$S_\text{lmr}$} with a $\tau_w$ of 0.354 achieves a relative improvement of 70\% on VOC-FT compared to \textit{SFDA} with a $\tau_w$ of 0.108.}

{
As summarized in Table \ref{tab:VOC}, \textit{ETran} outperforms all the previous works and baselines in terms of all the evaluation metrics. The results of \textit{ETran} only with the energy score (without labels) are also presented, which provide comparable or better scores than the previous classification-only scores.}

{\bf COCO.}
{
The comparison results for the COCO-based benchmark are provided in Table \ref{tab:coco}. Although adding the LogME's regression score to the previous improves their results, \textit{ETran} still achieves the best performance.}

{\bf HuggingFace.} The comparison results on HF are summarized in Table \ref{tab:HugFace}, which are very insightful for evaluating the performance of the transferability metrics when the source models have different architectures. Similar to VOC benchmark, the regression-empowered baselines outperform the previous classification-only works by an average of 0.047 in $\tau_w$ (average $\tau_w$ of 0.307 vs. 0.260). 
In overall, the proposed \textit{ETran} method achieves the best results in terms of all the evaluation metrics.
As presented in Table \ref{tab:HugFace}, we obtain larger numbers in terms of ${Pr(\text{top-k})}$ on the HF benchmark compared with the VOC and COCO benchmarks. We argue that when the source models have different architectures, the tranferbaility estimation is less challenging because the features extracted by the source models are more distinct. This is well-aligned with the {results} of the classification benchmark 
designed in the previous works in which the source models have different architectures \cite{sfda}.

\vspace{5pt}
\begin{table}[ht!]\setlength\tabcolsep{1.5pt}
\footnotesize
\centering
\caption{Ablation study of ETran scores on object detection benchmarks. 
}
\label{tab:Ablation_od}
\begin{tabular}{ccc|cc|cc|cc|cc}
\toprule
&&&\multicolumn{2}{c|}{VOC-FT}&\multicolumn{2}{c|}{VOC-RH}&\multicolumn{2}{c|}{COCO}&\multicolumn{2}{c}{HF}\\
 $S_\text{cls}$ & $S_\text{reg}$ &  $S_\text{en}$ & Pr(top3) & $\tau_{w}$ &Pr(top3) & $\tau_{w}$ &Pr(top3) & $\tau_{w}$&Pr(top3) & $\tau_{w}$ \\
\hline

\checkmark &           &             &0.29 & 0.14&0.43 & 0.37&0.53&0.13&0.80&0.38\\
& \checkmark&             &0.39 &0.36&0.50 &0.54 &0.40&0.12&0.80&0.51\\

&           & \checkmark  & 0.46&0.31&0.43&0.32&0.60&0.21&0.80&0.41 \\

\checkmark&\checkmark &             & 0.25& 0.31&0.64& 0.57&0.47&0.25&1.00& 0.50\\
    
&\checkmark &\checkmark   &\bf{0.50} & 0.40&0.57 & 0.55&0.53&0.23&0.80& 0.45  \\

\checkmark &           &\checkmark   &\bf{0.50} &0.37&0.50 & 0.41&0.53&0.32&0.80&0.40 \\

\checkmark &\checkmark &\checkmark   &\bf{0.50}&\textbf{0.44}&{\bf0.68}&\bf{0.59}&{\bf0.60}&{\bf0.33}&\bf{1.00}&\bf{0.52}\\
\bottomrule
\end{tabular}
\end{table}

\begin{table}[!htb]\setlength\tabcolsep{3pt}
    \caption{\textbf{Left}: Ablation study on ETran scores for classification benchmark. \textbf{Right}: Ablation study on the logits- vs. features-based energy scores.}
    \label{tab:Ablation_c}
    \begin{minipage}{.4\linewidth}
        \centering
        \footnotesize

        \begin{tabular}{cc|c}
        \toprule
         $S_\text{cls}$ &  $S_\text{en}$  & $\tau_{w}$ \\
        \hline

        \checkmark &           & 0.470 \\

        & \checkmark& 0.475\\

        \checkmark &\checkmark &{\bf0.562}\\
        \bottomrule
        \end{tabular}
    \end{minipage}%
    \begin{minipage}{.6\linewidth}
        \footnotesize
        \centering
        \begin{tabular}{ c|ccc}
        \toprule
        Method&\multicolumn{3}{c}{$\tau_w$}  \\
        & CB & VOC & HF \\
        \hline
        \textbf{Logits}&-0.088&0.270&0.297\\
        \textbf{Features}&\textbf{0.475}&\textbf{0.309}&\textbf{0.412}\\
        \bottomrule
        \end{tabular}
    \end{minipage} 
\end{table}

\begin{table}[h!]\setlength\tabcolsep{2pt}
    \caption{Time complexity analysis.} 
    \centering
    \footnotesize
    \begin{tabular}{lc|lc}
    \toprule
    \multicolumn{2}{c|}{Object Detection (VOC-FT)}&\multicolumn{2}{c}{Classification}\\
    Method&Run-Time (s)& Method&Run-Time(s) \\ \hline
    LogME \cite{logme} &{\bf11}& LogME \cite{logme} &{\bf65}\\
    PACTran \cite{pactran} &53& PACTran \cite{pactran} &444 \\
    SFDA \cite{sfda}&28 & SFDA \cite{sfda} &236\\ \hline
    LogME+$S_\text{lmr}$&22&LogME+$S_\text{en}$&70\\
    PACtran+$S_\text{lmr}$&63&PACTran+$S_\text{en}$&449\\
    SFDA+$S_\text{lmr}$ &39&SFDA+$S_\text{en}$&240 \\ \hline
     \textbf{ETran ($S_\text{en}$)}&0.3&\textbf{ETran ($S_\text{en}$)}&5 \\
    \textbf{ETran ($S_\text{cls}\text{+}S_\text{en}\text{+}S_\text{reg}$)} &\underline{25}& \textbf{ETran {($S_\text{cls}\text{+}S_\text{en}$)}} &\underline{101}\\
    \bottomrule

    \end{tabular}
    \label{tab:run-time}
\end{table}

\subsection{Ablation and Complexity Study}  

\textbf{Components of ETran.}
The individual performance of the proposed \textit{ETran}'s classification, energy, and regression scores on all the object detection and classification benchmarks is summarized in Tables \ref{tab:Ablation_od} and \ref{tab:Ablation_c}.

As presented in Table \ref{tab:Ablation_od}, excluding any of the scores from \textit{ETran} can result in performance drop for all the benchmarks. It is also shown that the regression score has the major contribution in the object detection task. This is an important finding that emphasizes the limitation of the previous classification-only transferability metrics for object detection task. On the other hand, Table \ref{tab:Ablation_c}-Left shows the importance of both classification and energy scores on the classification scenarios.

\textbf{Energy Score.}
{In Table \ref{tab:Ablation_c}-Right, the performance of the proposed energy-based transferability score calculated over logits vs. features is given. As shown by the results, the feature-based energy score calculation significantly performs better than the logit-based scenario. As also discussed in Section \ref{ssec:ETran}, it is mainly because the features extracted by the source models acquire more general information about the target dataset. On the other hand, logits are the outputs of the network's head that is specific to the labels of the source datasets used for training the source models. Specifically, in the case that the number of labels in the source and target datasets are significantly different, using logits for calculation of energy score can be misleading. 
}

{\bf Time Complexity.} {The computational complexity of \textit{ETran} compared with the previous works and the baselines over the VOC-FT and classification benchmarks are given in Table \ref{tab:run-time}. The numbers in the table are the running time of the whole transferability assessment averaged over the number of target datasets (i.e., 28 for VOC-FT and 11 for the classification benchmark).}
Among the previous works, LogME is the fastest metric. Our \textit{ETran} is the second model after LogME and it is faster that PACTran and SFDA. \textit{ETran} ($S_\text{en}$) is around 10$\times$ faster than LogME, which shows the efficiency of the energy score calculation in our method. Running time of PACTran depends on the number of iterations used for optimization (i.e., 100 iterations by default). 
SFDA has a self-challenging mechanism that requires applying Fisher Discriminate Analysis twice on all the samples. In contrast, \textit{ETran} only needs one round for LDA-based score calculation, which makes it 2$\times$ faster than SFDA. 

\section{Conclusion}
In this work, we proposed an energy-based transferability metric for classification and object detection. We introduced energy score as a fast and unlabeled transferability score that is used together with labeled classification and regression scores, which outperformed previous transferability metrics on object detection and classification scenarios. In terms of running time, \emph{ETran} is comparable with the previous works, while obtains better performance. In this work, we only showed the performance of \emph{ETran} for vision tasks. Future work should evaluate the performance of this method for other modalities such as language models. 

{\footnotesize
\bibliographystyle{ieee_fullname}
\bibliography{egbib}
}

\clearpage
 \section{Appendix}
 This Appendix provides further details about \emph{ETran} and demonstrates further experimental results. In Section \ref{sec:target_selection} and \ref{sec:language_models}, we  further evaluate \emph{ETran} on \textit{target selection} scenario and also language models. In Sections \ref{sec:analysis}-\ref{sec:svd_vs_logme}, we provide theoretical and experimental analysis of classification, regression, and energy scores of \emph{ETran}. In \ref{sec:gt_finetuning}, the fine-tuning procedure and the resulting ground-truth validation scores on all the benchmarks are provided and in \ref{sec:limitation} limitations and future work are discussed.

 \subsection{Target Selection Scenario}
\label{sec:target_selection}
\begin{figure}[t]
\begin{center}
   \includegraphics[width=1.0\linewidth]{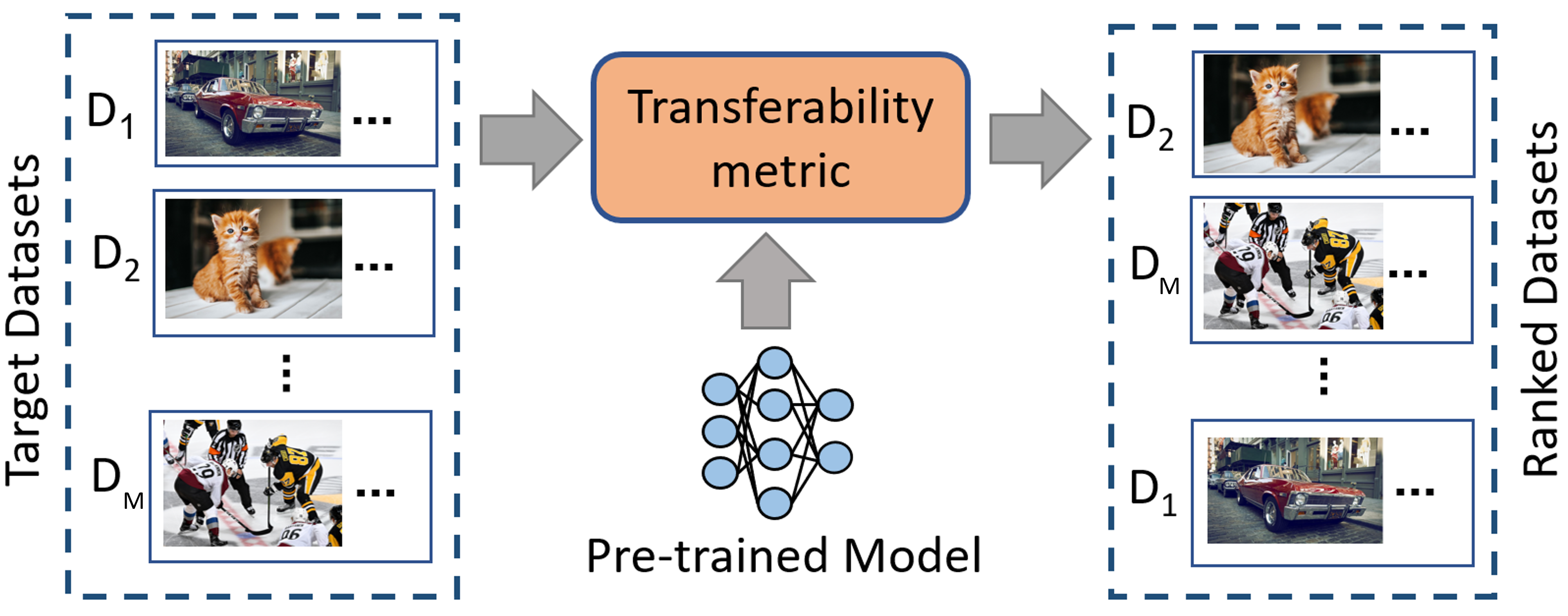}
\end{center}
    \caption{The overall framework of transferability estimation in the target selection scenario. Given $M$ target datasets and a source pre-trained model, the goal is
    to rank target datasets according to the actual performance of the source model after fine-tuning it on the target dataset. }
\label{fig:target_selection}
\end{figure}

In the main body of the paper, following most of the previous works, we considered that $M$ source models and a target dataset are given and the transferability metrics tried to rank the source models. Some of the previous works including GBC \cite{gbc} and LEEP \cite{LEEP} define a new scenario, where a single source model and multiple target datasets are given and the transferability metrics are used to rank the target datasets. We call this scenario as \textit{target selection}. Figure \ref{fig:target_selection} shows an overview of the \textit{target selection} scenario.

We use the classification benchmark \cite{sfda} explained in Section \ref{ssec:experiments_cls} to evaluate \emph{ETran}'s performance compared with the previous works on the target selection. Table \ref{tab:targt_selection} provides Kendall tau ({$\tau_w$}) of the 11 source classification models. \emph{ETran} obtains an average $\tau_w$ of 0.545 over 11 models, while SFDA, PACTran, LogME and LEEP obtain an average $\tau_w$ of 0.376, 0.295, -0.020, and 0.224, respectively. \emph{ETran} outperforms SFDA by a relative improvement of 45\% in-terms of {$\tau_w$}. Table \ref{tab:targt_selection} also shows that adding energy score (i.e., $S_\text{en}$) to the previous works improves their results on most of the cases (e.g., $\tau_w$ of 0.433 vs. 0.376 for SFDA). 

\subsection{Evaluation on Language Models}
\label{sec:language_models}
 Experiments on other modalities are our future work. In this section we show the preliminary results of our experiments on the language models, where we use RTE task from GLUE benchmark \cite{glue} and 10 popular pre-trained language models from HuggingFace (e.g., BERT, RoBERTa, BART, ALBERT, and DeBERTA). ETran, SFDA, and LogME obtain a $\tau_w$ of \textbf{0.421}, 0.391, and 0.138, respectively, which shows the superiority of ETran compared to others.

\begin{table*}\setlength\tabcolsep{4pt}
\footnotesize
\centering
\caption{The performance of \textit{ETran} compared with previous works for the target selection scenario on the classification benchmark (in terms of Kendall tau $\tau_w$).}
\label{tab:targt_selection}
\begin{tabular}{l|ccccccccccc|c}
	\toprule
	Method             &  Res34   &     Res50     &   Res101   &  Res152  & Dens169  & Dens121  &  Dens201      &MNas&Google &Inception   &Mobilenet & \textbf{Average}\\ \midrule
	LEEP \cite{LEEP}   &0.253&0.314&0.330&0.314&0.143&0.157&0.127&0.242&0.159&0.263&0.157&0.224\\ 
	LogME \cite{logme} &-0.387&-0.081&-0.118&-0.101&0.241&-0.226&0.207&0.203&-0.191&0.03&0.203&-0.020\\
	PACTran \cite{pactran} &0.373&0.467&0.402&0.397&0.260&0.295&0.047&0.243&0.214&0.394&0.154&0.295\\
	SFDA \cite{sfda} &0.501&0.501&0.484&0.501&0.301&0.314&0.284&0.312&0.211&0.462&0.269&0.376 \\
	\hline
	LEEP+$S_\text{en}$ &0.333&0.244&0.349&0.410&0.260&0.218&0.296&0.143&0.087&0.191&0.038&0.233\\
	LogME+$S_\text{en}$&-0.387&0.053&0.113&0.005&0.306&-0.211&0.302&0.161&-0.259&0.150&0.241&0.043\\
	PACTran+$S_\text{en}$ &0.350&0.496&0.445&0.392&0.241&0.295&0.106&0.201&0.229&0.239&0.209&0.291
 \\ 
	SFDA+$S_\text{en}$      &0.604&0.624&0.678&0.612&0.276&0.387&0.256&0.271&0.192&0.496&0.371&0.433 \\ \hline
	ETran ($S_\text{cls}\text{+}S_\text{en}$) &0.436&0.542&0.525&0.574&0.661&0.525&0.521&0.692&0.449&0.394&0.681&{\bf 0.545}\\
 \bottomrule
\end{tabular}
\end{table*}

\subsection{Analysis of \textit{ETran}'s Scores}
\label{sec:analysis}

In this section, we theoretically and experimentally analyze the proposed LDA-based classification and SVD-based regression scores compared with their peers including SFDA-based classification \cite{sfda} and LogME-based regression \cite{logme} scores, respectively. 

Before the analysis, we will first recap the intuition for transferability scores. The transferability score measures the compatibility between the extracted features and the ground-truth labels. 

\begin{table*}\setlength\tabcolsep{5pt}
\footnotesize
\centering
\vspace{5pt}
\caption{Comparing LDA and SFDA on the classification benchmark based on $\tau_w$. The self-challenging mechanism of SFDA diminishes the performance on many datasets.} 
\label{tab:lda-sfda-class}
\begin{tabular}{l|ccccccccccc|c}
	\toprule
	Method             &  CIFAR10   &     VOC     &   Caltech-101   &  AirCraft  & CIFAR100  & Food-101  &  Pets      &Flowers    &Cars &DTD   &Sun & \textbf{Average}\\ \midrule
	SFDA \cite{sfda} &  \bf{0.849}   &    0.518    &   \bf{0.555}   &   -0.215   & 0.793 & 0.427  & 0.340 & {\bf0.590} &{\bf0.312}&{\bf0.633}&{\bf0.722}&{\bf0.502} \\
 	LDA ($S_\text{cls}$) &   0.842    &   {\bf0.521}   &    0.354    &   {\bf-0.146}   & {\bf0.869}& {\bf0.754} & {\bf0.713} & 0.357 &-0.006&0.303&0.616& 0.470\\
	\hline
	SFDA + $S_\text{en}$      &    {\bf 0.890}    &    0.606    &   {\bf 0.558}    &   -0.161   &   0.856    &  0.370     &   0.422    &0.406  &{\bf 0.328}&{\bf0.639}&{\bf0.744}&0.514 \\
	LDA ($S_\text{cls}$) + $S_\text{en}$       &   0.887    &   \textbf{0.667}    &    0.440     &   {\bf-0.091}   &   {\bf 0.900}   &  {\bf0.829}     &  {\bf0.713}     & {\bf0.580} &0.246&0.303&0.708&\textbf{0.562} \\
 \bottomrule
\end{tabular}
\end{table*}

\begin{table}\setlength\tabcolsep{4pt}
    \footnotesize
    \centering
    \caption{Comparing LDA and SFDA on object detection benchmarks. 
    }
    \begin{tabular}{l|cc|cc|cc}
    \toprule
        &\multicolumn{2}{c|}{VOC-FT} & \multicolumn{2}{c|}{COCO}  & \multicolumn{2}{c}{HF}\\ 
        &Pr(top3)&$\tau_w$&Pr(top3)&$\tau_w$&Pr(top3)&$\tau_w$\\ \hline
        SFDA \cite{sfda}&0.250&0.108&\bf{0.533}&0.104&\bf{1.000}&0.312\\
        LDA (ours)&\bf{0.286} &\bf{0.141} &\bf{0.533}&\bf{0.131}&0.800&\bf{0.376}\\
    \bottomrule
    \end{tabular}
    \label{tab:lda-sfda}
\end{table}

More formally, each sample in the target dataset comes from an underlying distribution $\mathzapf{D}$.
To avoid costly fine-tuning, it is assumed that the source model's backbone is frozen. 
The extracted feature ${f}$ and its corresponding labels $y$ come from a feature distribution
$\mathzapf{F}$, denoting as $(f, y) \sim  \mathzapf{F}$. 

For classification, $y$ is a scalar for the target (ground-truth) class. For regression (specifically for the task of object detection), $y$ means the position and scale for bboxes. Since it is natural that separate weights are leveraged for predicting each component of position and scale, they can be treated independently for transferability score.
Thus, for simplicity, $y$ here is a scalar for one component of position and scale.  

From distribution $\mathzapf{F}$, we have $K$ samples (i.e., bboxes for object detection) and their corresponding labels, i.e., $(f, {y}) \sim \mathzapf{F}^K$. 
Here,  $f \in \mathbb{R}^{K\times \hat{h}}$ is the extracted feature matrix. 
 ${y} \in \mathbb{R}^{K}$ is the labels of samples. 
Given $f$ and $y$, the transferability score measures the compatibility between the feature and label. 
We say they are compatible if there exists a mapping from feature space to  label space and this mapping is accurate for the feature and label pair from $\mathzapf{F}$. To conclude, transferability score of a source model  towards $\mathzapf{D}$ is measured by the generalization performance of the mapping on $\mathzapf{F}$. 

There are two challenges: 1) After fine-tuning the source model, feature distribution $\mathzapf{F}$ drift. The transferability score fail to compensate for this. 
2) The generalization performance is defined on distribution $\mathzapf{F}$, however, only $K$ samples from distribution are available. Thus, it matters whether the estimation of transferability score is tight or vacuous. Motivated by these two challenges, LDA-based classification   and SVD-based regression scores are proposed. We will give detailed analysis in the following sections. 

Note that for the simplicity and generality of our method on different benchmarks, the three transferability scores in Eq. \ref{eq:overall_score} are normalized between [0,1] and equally summed. Based on our initial study, having different weights for the normalized three terms does not significantly affect the final results. It is a common practice to use fixed hyperparameters for generality (i.e., PACTran \cite{pactran}).

\subsection{Energy Score vs. Classification Score.} 
\label{sec:energy_vs_cls}
As studied in \cite{elang,oodenergy}, the softmax score for a classifier, $\Phi$ with $C$ classes, is defined as:
\vspace{-5pt}
\begin{equation}
\underset{y}{\text{max }} p(y|x)=  \underset{y}{\text{max }}\frac{e^{\Phi^{(y)}(x)}}{\sum_c^{C}e^{\Phi^{(c)}(x)}}=\frac{e^{\Phi_{max}(x)}}{\sum_c^{C}e^{\Phi^{(c)}(x)}}.   
\end{equation}
If we take the logarithm of both sides we have: 

\vspace{-10pt}
\begin{align}
 \log\underset{y}{\text{max }} p(y|x) &= \Phi_{\text{max}}(x)-\log\sum_{c}^{C}e^{\Phi^{(c)}(x)} \nonumber \\
 &= \Phi_{\text{max}}(x) + E(x). \label{eq:took-log}
\end{align}

Therefore, the log of softmax confidence score is in fact energy score shifted with the maximum value of logits. Since $\Phi_{max}(x)$ tends to be higher and $E(x)$ (Eq. \ref{eq:energy_f1} in the paper) tends to be lower for in-distribution samples, the softmax confidence score is a biased scoring function that is no longer proportional to the probability density $p(x)$. Having $E(x)$ from Eq. \ref{eq:energy_constz} in the paper, we can write Eq.{~\ref{eq:took-log}} as: 
\vspace{-5pt}
\begin{equation}
    \log\underset{y}{\text{max }} p(y|x)=-\log p(x)+\underbrace{\Phi_{max}(x) - \log Z}_{\text{not constant, larger for in-dist x}}.
\end{equation}
Thus, unlike the energy score (as proved in Section \ref{ssec:ETran} of the paper), the softmax classification score is not well-aligned with $p(x)$ \cite{elang,oodenergy}, which makes it less reliable for out-of-distribution detection and transferability assessment.

\subsection{LDA-Based Classification Score vs. SFDA}
\label{sec:lda_vs_cls}

The feature distribution $\mathzapf{F}$ shifts from source to target dataset after fine-tuning. The features $f$ extracted by the pre-trained models are separable based on the source dataset’s classes. However, after fine-tuning, the features are separable based on the target dataset’s classes. To mitigate feature distribution shift, we propose to use LDA-based classification score. Linear discriminant analysis (LDA) projects the features into a space that is separable w.r.t the target classes. This coincides with the feature from fine-tuned model and thus mitigate distribution shift.

{
Compared to our LDA-based score, SFDA \cite{sfda} has a self-challenging mechanism, which has two drawbacks:
1) The practical computational cost of SFDA is almost twice LDA, which is because the self-challenging mechanism performs Linear Discriminate Analysis twice on all the samples. 2) The self-challenging mechanism also introduces noise on the features, which can negatively affect the deep connection between the discriminative and energy-based models, i.e., the linear alignment of the calculated negative free energy with the likelihood function (as discussed in Section \ref{ssec:ETran}).}
 
{
As summarized in Table \ref{tab:lda-sfda-class}, SFDA overally performs a bit better than the LDA-based score (i.e., $S_\text{cls}$) on the classification benchmark. However, when integrated with the proposed energy score (i.e., $S_\text{en}$), LDA archives a better performance. Table \ref{tab:lda-sfda} also compares the performance of LDA vs. SFDA on the object detection benchmarks, which shows that LDA outperforms SFDA on three benchmarks in terms of $\tau_w$.
}

\begin{figure}
    \centering
    \includegraphics[width=1\linewidth]{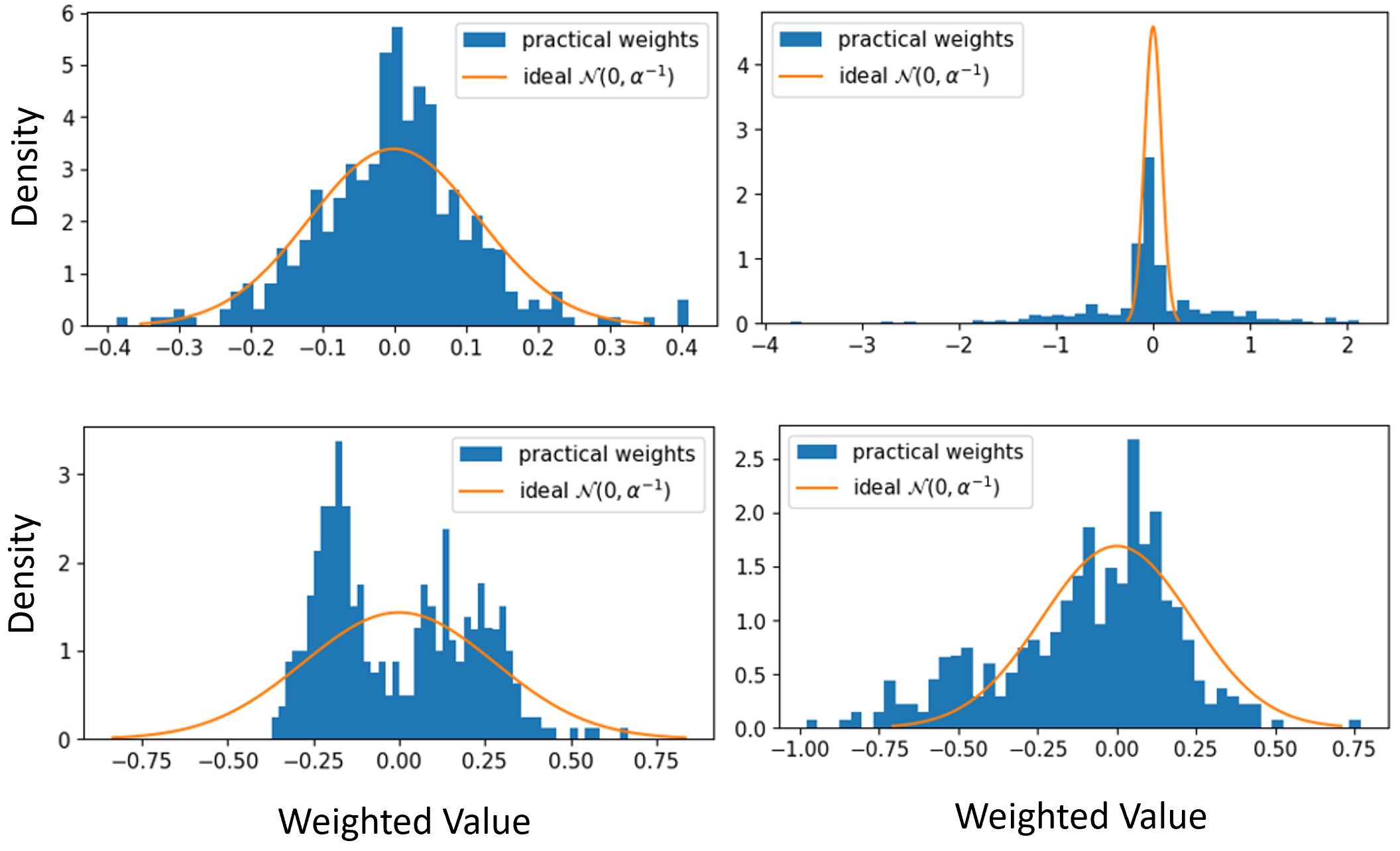}
    \caption{LogME's assumption analysis. \textbf{Blue}: the histogram of practical weights. \textbf{Orange}: the weights distribution with optimal $\alpha$. Weights distribution comes from different pairs of source models and target datasets in the VOC-FT benchmark. } 
    \label{fig:logme-assumption}
\end{figure}

\begin{table}[t]\setlength\tabcolsep{3.2pt}
    \footnotesize
    \centering
    \vspace{10pt}
    \caption{Comparing LogME and SVD-based regression scores on object detection benchmarks.}
    \begin{tabular}{l|cc|cc|cc}
    \toprule
        &\multicolumn{2}{c|}{VOC-FT} & \multicolumn{2}{c|}{COCO}  & \multicolumn{2}{c}{HF}\\ 
        &Pr(top3)&$\tau_w$&Pr(top3)&$\tau_w$&Pr(top3)&$\tau_w$\\ \hline
        LogME ($S_{\text{lmr}}$) \cite{logme}& 0.357& 0.356& 0.400&0.113&0.800&0.400 \\
        SVD-reg ($S_{\text{reg}}$)& \bf{0.393}& \bf{0.357}&0.400 &\bf{0.122}&0.800&\bf{0.512} \\
    \bottomrule
    \end{tabular}
    \label{tab:log-svd}
\end{table}

\begin{table*}\setlength\tabcolsep{4.8pt}
\footnotesize
\centering
\caption{The fine-tuning accuracy (map50) of pre-trained models on VOC-FT benchmark. The best and the second best pre-trained source models for a given target dataset are shown with bold and underline, respectively.} 
\label{tab:gt_voc_ft}
\begin{tabular}{ll|ccccccccccccccccccc}
    \toprule
    &&\multicolumn{19}{c}{Source Models} \\
    &&1&2&3&4&5&6&7&8&9&10&11&12&13&14&15&16&17&18&19\\ \hline
    \multirow{28}{*}{\rotatebox[origin=c]{90}{Target Datasets}}&1&0.28&0.33&0.31&\underline{0.36}&0.34&0.33&0.35&0.33&0.29&0.33&0.35&0.32&0.35&0.33&0.32&0.32&0.33&\bf{0.36}&0.32\\
    &2&0.30&0.33&0.33&0.36&0.34&0.33&0.34&0.33&0.29&0.31&{\bf0.38}&0.33&0.34&0.33&0.34&0.34&0.34&\underline{0.37}&0.32\\
    &3&0.38&0.46&0.43&{\bf0.52}&0.46&0.47&0.49&0.47&0.44&0.42&0.49&0.46&0.47&0.47&0.47&0.43&0.48&\underline{0.50}&0.46\\
    &4&0.33&0.36&0.39&\underline{0.41}&0.37&0.38&0.40&0.40&0.35&0.33&0.41&0.36&0.40&0.40&0.38&0.36&0.37&{\bf0.42}&0.37\\
    &5&0.47&0.50&0.48&\underline{0.53}&0.51&0.50&0.50&0.48&0.49&0.47&0.51&0.48&0.51&0.49&0.49&0.49&0.49&\bf{0.54}&0.51\\
    &6&0.48&0.49&0.49&\bf{0.55}&0.48&0.51&0.52&0.51&0.48&0.46&0.51&0.49&0.51&0.51&0.51&0.46&0.51&\underline{0.52}&0.49\\
    &7&0.45&0.45&0.46&0.51&0.49&0.48&0.49&0.46&0.44&0.44&\bf{0.53}&0.45&0.48&0.50&0.51&0.45&0.51&\underline{0.51}&0.47\\
    &8&0.24&0.31&0.29&0.33&0.31&0.32&0.32&0.29&0.24&0.28&\bf{0.33}&0.29&0.33&0.31&0.31&0.32&0.33&\underline{0.33}&0.33\\
    &9&0.41&0.47&0.46&0.48&0.48&\underline{0.50}&0.48&0.45&0.40&0.42&0.47&0.42&0.48&0.45&0.49&0.46&0.47&\bf{0.50}&0.47\\
    &10&0.27&0.32&0.35&0.36&0.34&0.34&\bf{0.37}&\bf{0.37}&0.28&0.32&0.35&0.32&0.35&0.32&0.35&0.32&0.34&\underline{0.36}&0.33\\
    &11&0.46&0.49&0.50&0.51&0.49&0.48&0.47&0.51&0.47&0.49&0.51&0.46&0.49&0.49&0.50&\underline{0.51}&0.49&\bf{0.53}&0.49\\
    &12&0.44&0.51&0.49&0.49&0.50&\underline{0.51}&0.50&0.50&0.45&0.48&0.50&0.45&0.50&0.49&0.50&0.48&0.49&\bf{0.52}&0.51\\
    &13&0.38&0.44&0.47&\bf{0.49}&0.44&0.44&0.44&0.47&0.38&0.41&0.47&0.44&0.46&0.43&0.46&0.42&\underline{0.47}&0.46&0.44\\
    &14&0.42&0.46&0.43&0.47&\underline{0.47}&\underline{0.47}&0.46&0.47&0.41&0.40&\bf{0.49}&0.43&0.47&0.47&\bf{0.49}&0.44&0.47&0.47&0.47\\
    &15&0.30&0.34&0.34&0.35&0.35&0.33&0.34&0.33&0.29&0.32&\underline{0.37}&0.31&0.33&0.33&0.35&0.33&0.35&\bf{0.37}&0.34\\
    &16&0.50&0.51&0.51&\underline{0.52}&0.52&0.50&0.49&0.50&0.49&0.51&0.51&0.49&0.51&0.50&0.51&0.51&0.51&\bf{0.53}&0.52\\
    &17&0.45&\underline{0.51}&0.48&0.51&0.49&0.49&0.49&0.49&0.45&0.48&\bf{0.51}&0.46&0.51&0.50&0.50&0.48&0.49&0.50&0.50\\
    &18&0.37&0.41&0.41&\bf{0.44}&0.42&0.43&0.39&0.41&0.36&0.37&0.43&0.41&0.42&0.43&0.43&0.40&0.43&\underline{0.43}&0.41\\
    &19&0.50&0.52&0.51&0.54&0.53&0.57&0.53&0.54&0.50&0.50&0.56&0.51&0.53&0.53&\bf{0.57}&0.49&0.50&\underline{0.56}&0.56\\
    &20&0.59&0.61&0.60&\underline{0.63}&0.62&0.63&0.62&0.61&0.59&0.61&0.60&0.57&0.62&0.59&0.61&0.60&0.61&\bf{0.64}&0.61\\
    &21&0.49&0.55&0.49&0.51&0.50&0.54&0.53&0.52&0.47&0.47&\bf{0.56}&0.49&0.54&0.53&0.48&0.47&0.49&\underline{0.55}&0.50\\
    &22&0.55&0.61&0.56&0.62&0.60&0.63&0.61&0.61&0.54&0.56&\underline{0.63}&0.60&0.63&0.60&\bf{0.64}&0.59&0.59&0.61&0.61\\
    &23&0.51&0.53&0.52&\bf{0.55}&0.53&0.53&0.53&0.52&0.51&0.51&0.54&0.49&0.52&0.52&0.53&0.54&0.52&\underline{0.54}&0.52\\
    &24&0.55&0.54&0.55&0.60&0.55&0.58&0.58&0.56&0.55&0.55&0.59&0.53&0.59&0.57&\bf{0.60}&0.53&0.54&0.58&\underline{0.60}\\
    &25&0.47&0.44&0.47&0.54&0.47&0.50&0.51&0.51&0.42&0.46&\underline{0.53}&0.46&0.52&0.48&0.52&0.46&0.51&\bf{0.55}&0.51\\
    &26&0.65&0.68&0.66&\bf{0.69}&0.67&0.68&0.68&0.67&0.66&0.66&0.68&0.67&\bf{0.69}&0.67&0.67&0.68&0.67&0.68&\underline{0.69}\\
    &27&0.57&0.14&0.56&\bf{0.60}&0.58&0.58&0.58&0.58&0.55&0.57&0.59&0.54&0.57&0.58&\underline{0.60}&0.58&0.59&0.58&0.57\\
    &28&0.58&0.64&0.62&\bf{0.69}&   0.63&0.65&0.66&0.65&0.60&0.61&0.68&0.60&0.66&0.63&0.66&0.62&0.64&\underline{0.68}&0.63\\

 \bottomrule
\end{tabular}
\end{table*}

\begin{table*}\setlength\tabcolsep{6pt}
\footnotesize
\centering
\vspace{5pt}
\caption{The fine-tuning accuracy (map50) of pre-trained models on COCO benchmark. The best and the second best pre-trained source models for a given target dataset are shown with bold and underline, respectively.} 
\label{tab:gt_coco}
\begin{tabular}{ll|ccccccccccccccc}
    \toprule
    &&\multicolumn{15}{c}{Target Datasets} \\
    &&1&2&3&4&5&6&7&8&9&10&11&12&13&14&15\\ \hline
    \multirow{9}{*}{\rotatebox[origin=c]{90}{Source Models}}&1&0.344&0.445&0.281&0.177&0.593&0.524&\bf{0.608}&\underline{0.425}&\bf{0.342}&0.263&\bf{0.430}&0.141&0.305&0.440&0.319\\
    &2&\bf{0.373}&0.489&\bf{0.304}&\underline{0.194}&0.632&\bf{0.556}&0.588&\bf{0.432}&\underline{0.334}&\underline{0.277}&\underline{0.415}&\bf{0.183}&\underline{0.343}&0.458&0.344\\
    &3&0.340&0.481&0.311&0.171&0.573&0.534&\underline{0.559}&0.412&0.334&0.249&0.403&0.171&0.306&0.452&0.354\\
    &4&0.353&0.482&0.278&0.172&0.614&0.523&0.533&0.406&0.311&0.252&0.412&0.161&\bf{0.346}&0.460&0.385\\
    &5&0.338&0.462&0.283&0.186&0.600&0.543&0.580&0.385&0.310&0.237&0.392&0.159&0.328&0.450&0.352\\
    &6&0.355&\bf{0.512}&\bf{0.304}&0.189&\bf{0.638}&\underline{0.547}&0.526&0.419&0.310&0.259&0.413&\underline{0.181}&0.322&0.464&\underline{0.358}\\
    &7&0.353&0.479&0.324&\bf{0.202}&\underline{0.636}&0.515&0.585&0.407&0.330&0.255&0.412&0.173&0.327&\underline{0.473}&\bf{0.389}\\
    &8&0.355&\underline{0.493}&\underline{0.300}&0.181&0.635&0.535&0.555&0.424&0.319&\bf{0.303}&0.422&0.158&0.330&\bf{0.484}&0.384\\
    &9&\underline{0.365}&0.471&0.290&0.186&0.616&0.524&0.548&0.402&0.333&0.248&0.400&0.157&0.400&0.449&0.364\\
 \bottomrule
\end{tabular}
\end{table*}

\begin{table}\setlength\tabcolsep{5pt}
\footnotesize
\centering
\caption{The fine-tuning accuracy (map50) of pre-trained models on HuggingFace benchmark. The best and the second best pre-trained source models for a given target dataset are shown with bold and underline, respectively.} 
\label{tab:hugface}
\begin{tabular}{l|ccccc}
	\toprule
	&NFL&Blood&CSGO&Forklift&Valorant\\ \hline
	Yolov5s \cite{yolov5}&0.261&0.902&\underline{0.924}&0.838&\underline{0.982}\\
	Yolov5m \cite{yolov5}&\bf{0.314}&0.905&\bf{0.932}&\bf{0.852}&\bf{0.990}\\
	Yolov5n \cite{yolov5}&0.217&\underline{0.923}&0.908&0.789&0.959\\
	Yolov8s \cite{yolov8}&0.279&0.917&0.886&\underline{0.851}&0.971\\
	Yolov8m \cite{yolov8}&\underline{0.287}&\bf{0.927}&0.892&0.846&0.965\\
	Yolov8n \cite{yolov8}&0.209&0.893&0.844&0.838&0.937\\
 \bottomrule
\end{tabular}
\end{table}

\subsection{SVD-Based Regression Score vs. LogME}
\label{sec:svd_vs_logme}
We will first recap LogME score, analyze its problem, and then propose our solution.
{
LogME assumes that the weights of a linear model that maps from feature space to label space $\hat{y}=\mathbf{w}^\top {f}$, have a normal distribution as follows:  $\mathbf{w} \sim \mathzapf{N}(0,\alpha^{-1}I)$. 
The prior of weights, $\alpha$, is optimized on target features, $f$. Then, the log marginal likelihood (i.e., evidence) of observing $f$ given ground-truth labels $y$ is used to measure the generalization performance of the pre-trained source models \cite{logme}.
}

If the assumption $\mathbf{w} \sim \mathzapf{N}(0,\alpha^{-1}I)$ matches the actual feature data, LogME score measures the generalization performance tightly. But if not, LogME will deviate from actual performance. 
{
In practice, there are many cases where the assumption of LogME does not hold.
}

{
In order to evaluate the assumption of LogME,
}
we calculate the optimal $\alpha$ using LogME and compare $\mathzapf{N}(0, \alpha^{-1} I)$ with the practical weight distribution of the last-layer weights of the fine-tuned models in the VOC-FT benchmark. The comparison is illustrated in Figure~\ref{fig:logme-assumption}. The first row in Figure~\ref{fig:logme-assumption} shows the successful cases that the practical weights follows Gaussian, where LogME finds the optimal $\alpha$. However, as shown in second row, there are cases that the practical weights cannot be described by Gaussian distribution. To this end, if the model's hypothesis space is limited, the derived transferability metric will be a vacuous bound of the model's actual performance.

{
In our SVD-based regression method,}
we relax the assumption, resort to SVD-based linear regression, and derive the transferability metric. We verified that this strategy is effective in practice. This simple strategy first finds the optimal mapping by solving best $\mathbf w^*$ for $ \arg\min_{\mathbf w} \|y-f \mathbf w\|^2$ and then measures the performance by negative remaining error $-\| y - f \mathbf w^* \|$. To prevent overfitting, more advanced way is to split $f$ into train and test sets by 7:3 ratio and evaluate performance on the test set. 

Considering $f$ is near rank-deficient and may be ill-conditioned (the bottom-level singular value is close to zero), we apply truncated SVD to obtain $\mathbf w^*$.  With SVD decomposition $f=\mathbf U \text{diag}(\mathbf{s}) \mathbf V^\top$,  $\mathbf w^*=\mathbf V  \text{diag}(\hat{\mathbf{s}})^{-1}\mathbf U^\top$, where $\text{diag}({\hat{\mathbf{s}}})$ is the truncated singular values whose top 80\% is preserved. 
With SVD, we approximately solve the linear regression in a way that it is less sensitive to errors and more effective for ill-conditioned matrices. 
Moreover, the complexity of our proposed regression score is $\mathcal{O}(n\hat{h}^2)$.

It is more efficient compared to LogME's $\mathcal{O}\left(n\hat{h}^2+\hat{h}^3+t(\hat{h}^2+n\hat{h})\right)$, where $t$ is the iteration for LogME to converge and $n$ is number of samples in the dataset.
Since the tailing singular value is truncated, the practical run-time of our proposed score will be further reduced. 

{
Table \ref{tab:log-svd} shows the performance of LogME (i.e., $S_{\text{lmr}}$) vs. our SVD-based regression (i.e., $S_{\text{reg}}$) on the object detection benchmarks. As seen by the results, the proposed SVD-based regression outperforms LogME on all the three benchmarks in terms of $\tau_w$ and $Pr(top3)$.
}

\subsection{Fine-Tuning and Ground-Truth Results}
\label{sec:gt_finetuning}
The ground-truth ranking of the pre-trained source models is obtained by fine-tuning each of the source models on all the target datasets. In this section, we provide the details of fine-tuning procedure for object detection and classification benchmarks. 

\textbf{VOC and COCO.} For the VOC and COCO benchmarks, we first train Yolov5s \cite{yolov5} on each of the source models for 300 epochs. The pre-trained source models are then fine-tuned on the train-set of the target datasets for 60 epochs. Tables \ref{tab:gt_voc_ft} and \ref{tab:gt_coco} show the map50 of the fine-tuned models on the validation-set of the target datasets. The best and second best pre-trained source models for a given target dataset are shown with bold and underline, respectively.

\textbf{HuggingFace.} We use 6 object detection models including: YOLOv5s, YOLO5m, YOLOv5n \cite{yolov5}, YOLOv8s, YOLOv8m, and YOLOv8n \cite{yolov8}. All the models were pre-trained on COCO dataset using the default setting \cite{yolov5,yolov8}. 
Table \ref{tab:hugface} provides the map50 of the source models after fine-tuning on the target datasets.

\textbf{Classification.}
The source models were pre-trained on ImageNet and were downloaded from Pytorch repository. The accuracy of the models after fine-tuning on the target datasets were obtained from SFDA \cite{sfda}. SFDA \cite{sfda} provides details of fine-tuning on each of the target datasets. Table 6 in the appendix of SFDA paper shows the accuracy of the pre-trained models after fine-tuning on each of the target datasets. 

\subsection{Limitations}
\label{sec:limitation}
We have briefly discussed the limitations in Section \ref{ssec:experiments_cls} of the main body. 
Figure \ref{fig:energy_bad} of the main body shows two failure cases where the energy score does not always correlate positively with the accuracy of the target dataset.
In this section we further discuss the limitations of our work that need to be addressed in future work including: 
\textbf{1)} In all 4 benchmarks, source models differ either in their architectures or source datasets. 
It will be comprehensive to further validate considering both. 

\textbf{2)} The stability of our method \textit{w.r.t} the small perturbation on the target dataset and source pre-trained models needs to be studied further. \textbf{3)} In all scenarios, both source and target tasks were identical, e.g., both aimed for classification tasks. The stability of the method, when source and target tasks differ, should be investigated. \textbf{4)} Experiments on other modalities such as language models.

\end{document}